# USING LINGUISTIC ANALYSIS TO TRANSLATE ARABIC NATURAL LANGUAGE QUERIES TO SPARQL

Iyad AlAgha

Faculty of Information Technology, The Islamic University of Gaza, Gaza Strip, Palestine

`ialagha@iugaza.edu.ps`

## ABSTRACT

*The logic-based machine-understandable framework of the Semantic Web often challenges naive users when they try to query ontology-based knowledge bases. Existing research efforts have approached this problem by introducing Natural Language (NL) interfaces to ontologies. These NL interfaces have the ability to construct SPARQL queries based on NL user queries. However, most efforts were restricted to queries expressed in English, and they often benefited from the advancement of English NLP tools. However, little research has been done to support querying the Arabic content on the Semantic Web by using NL queries. This paper presents a domain-independent approach to translate Arabic NL queries to SPARQL by leveraging linguistic analysis. Based on a special consideration on Noun Phrases (NPs), our approach uses a language parser to extract NPs and the relations from Arabic parse trees and match them to the underlying ontology. It then utilizes knowledge in the ontology to group NPs into triple-based representations. A SPARQL query is finally generated by extracting targets and modifiers, and interpreting them into SPARQL. The interpretation of advanced semantic features including negation, conjunctive and disjunctive modifiers is also supported. The approach was evaluated by using two datasets consisting of OWL test data and queries, and the obtained results have confirmed its feasibility to translate Arabic NL queries to SPARQL.*

## 1. INTRODUCTION

The Semantic Web has emerged as an extension of the current Web, in which Web content has well-defined meaning through the provision of ontologies and machine-interpretable metadata. In recent years, a huge amount of data has been made available on the Web in RDF and OWL formats. However, current techniques for information retrieval from this semantic data restrict their use to only experienced users who have the ability to command formal logic. To allow ordinary users to interact with the Semantic Web content, several efforts have proposed natural language (NL) interfaces to ontologies and semantic knowledge bases [1]. These NL interfaces enable users to query ontologies and RDF stores by typing queries expressed in natural language. They provide approaches to translate NL queries to SPARQL, the formal query of the Semantic Web. Thus, they hide the formality of the semantic data as well as the executable query language.

Despite the considerable research that has explored NL interfaces to ontologies and RDF data, most efforts were designed to work with English. These efforts have benefited a lot from the advancement in the NLP of English and Latin based languages. However, there is very little, if any, attempt to support querying the Arabic content on the Semantic Web by using NL queries expressed in Arabic. This has been challenged by the difficulties associated with the Arabic NLP and the lack of efficient NLP tools similar to those available for the English language.

Arabic is the language spoken by hundreds of millions of people in Middle East and northern African countries, and is the religious language of all Muslims of various ethnicities around the world [2, 3]. There are various studies conducted by many that link the Arabic and Semantic values [4]. These studies have varied from the development of Arabic ontologies to the development of information retrieval and search systems. However, most of these studies were tailored to specific application needs, and often ignored the need to query the semantic content by using NL queries. On the other hand, some efforts have presented approaches for Arabic Question Answering (QA). However, they often did not consider the use of ontologies and semantic inference, and thus were not compatible with the data formats on the Semantic Web [5]. We believe that by enabling QA from the Arabic semantic content, we can make a step towards expanding the influence of ontologies and the Semantic Web among the Arab community.

To enable Arab users to query ontologies without being exposed to the underlying complexities, we propose a generic approach to translate Arabic NL queries to SPARQL. The proposed approach utilizes the off-the-shelf Arabic language toolkit [6] to build the parse tree of the user query. It then analyses the syntactic structure of the tree to extract NPs, identify head and modifier words and represent the query in a triple format, i.e. subject-predicate-object. The proposed approach is portable in terms that it can be easily ported from one ontology to another without significant effort.

## 2. RELATED WORK

Some efforts were conducted to build QA systems oriented to the Arabic language. Arabic QA applications can be divided into two types according to the covered domain of knowledge [5]:1) restricted domain QA systems which handle domain specific user queries. Examples of this type include AQAS [7] and QARAB [8]. 2) open domain QA systems which retrieve answers from heterogeneous databases such as the Internet. Common examples of this type are AQuASys [9] and IDRAAQ [10]. These efforts were limited to keyword-based search in raw documents. Answers were retrieved in the form of passages or documents where the concentration was on the morphological and syntactic aspects of Arabic. They do not handle ontology-based content or use deep reasoning for making sense of, and answering, user queries. Thus, they are not adequate for the Semantic Web use where data is published in RDF and OWL formats.

The support for Arabic language on the Semantic Web is still limited despite the considerable attention it has gained in the past few years. This can be attributed to the lack of ontologies expressed in Arabic and the complexities associated with the NLP of Arabic text [3]. In general, there are four different categories of research concerning the Arabic language and the Semantic Web [11]: 1) the development of Arabic ontologies [12-14], 2) Using ontologies for improving Arabic named entities extraction [15, 16], 3) Ontology based modelling of Islamic knowledge [17-19], and 4) supporting cross-language information retrieval [20, 21]. In parallel with these efforts, little attention was paid to enable Arab users to query Arabic ontologies through NL interfaces.

Several approaches adapted Information Retrieval (IR) approaches for making use of Arabic ontologies [22-24]. These approaches annotate Arabic documents using background domain ontologies. The search process is carried out by mapping the user keywords onto their semantic document annotations. These approaches are often not able to retrieve concise answers to questions but only a set of relevant documents or passages. In contrast, our work presents a generic and domain-independent approach for generating SPARQL from Arabic NL queries. From another perspective, the proposed approach can be used as an extension to Arabic ontology-based IR systems by supporting QA using NL queries rather than using traditional keyword-based search.

With respect to English language, several QA systems have been proposed. The input to these systems is generally a natural language query and the output is a list of relevant entities. These systems use two different approaches[25]: 1) Using linguistic approaches to capture complete triple-based patterns, including the relations, from the user query and match them to the underlying ontology [26-28]. 2) capturing ontology terms in the user query and then discovering relations between these terms from the knowledge base [29, 30]. Our approach falls in the first category as it adopts a linguistics-based approach, but focuses strictly on Arabic NL queries.

Common examples of linguistics-based approaches for interpreting NL queries to SPARQL include PowerAqua [26], PANTO [31] and Pythia [32]. PowerAqua can automatically query information from multiple ontologies at runtime. However, it lacks a deep analysis of language dependencies, and thus cannot handle complex queries. PANTO uses a statistical parser to build parse trees of NL queries and capture nominal phrase constituents. It then adopts a triple-based model to link and transform nominal phrases to SPARQL. Our approach was inspired from PANTO, but it was tailored to handle Arabic NL queries. Pythia is a QA system that also employs deep linguistic analysis. It can handle linguistically complex questions, but is highly dependent on a manually created lexicon. Therefore, it fails with datasets for which the lexicon was not designed.

The growing research on NL interfaces to ontologies has largely benefited from the advances of English NLP. However, there has not been a similar progress to support Arabic NLP, and the available NLP tools for Arabic are often imprecise and error-prone as compared to NLP tools for English [3]. The unique characteristics of Arabic language and its complex morphology make existing NL interfaces for the English text inefficient for Arabic.

## 3. A SAMPLE DOMAIN OF KNOWLEDGE

Before explaining the approach for translating Arabic queries to SPARQL, we introduce the sample ontology we built for illustration and testing purposes. Figure 1 depicts an excerpt of the ontology showing some ontology classes (e.g. Cure, Disease, Symptom, Organ and Diagnosis) as well as the relations between them, i.e. the object properties. Examples given in the paper use the schema of this ontology.

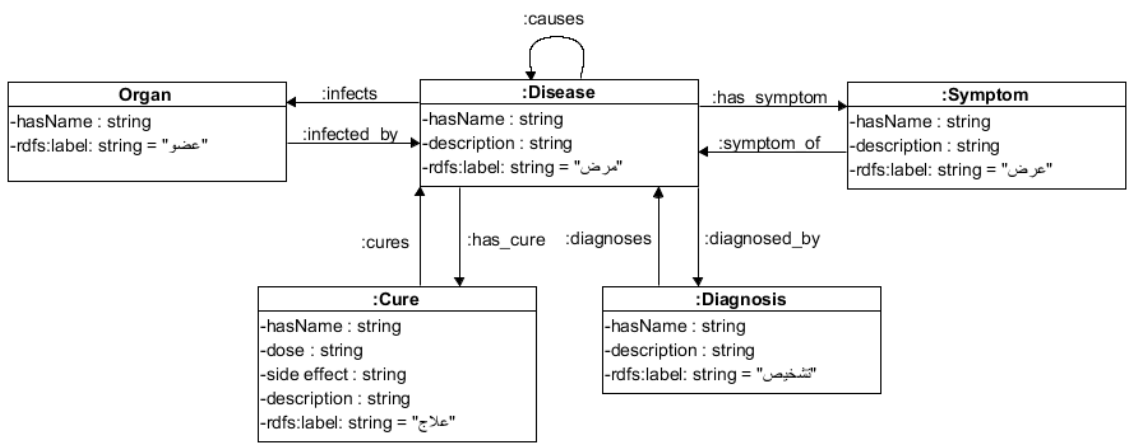

Figure 1. An excerpt of the Diseases Ontology

The translation from Arabic query to SPARQL requires mapping the Arabic words to the ontology terms that best describe them. To make the mapping of Arabic script possible, the ontology content should be either written in Arabic, or written in English but associated with

Arabic translations. For simplicity, we used the latter option by building the ontology in English and associating Arabic translations to the ontology terms by using the rdfs:label property. Therefore, the Arabic name of an ontology term can retrieved by reading the value of its rdfs:label property.

The approach for translating Arabic NL queries to SPARQL is shown in Figure 2. The input to the approach is the user query expressed in Arabic and the ontology representing the domain targeted by the query. The output is the SPARQL query that corresponds to the NL query. The steps of the proposed approach are explained in detail in the following sections.

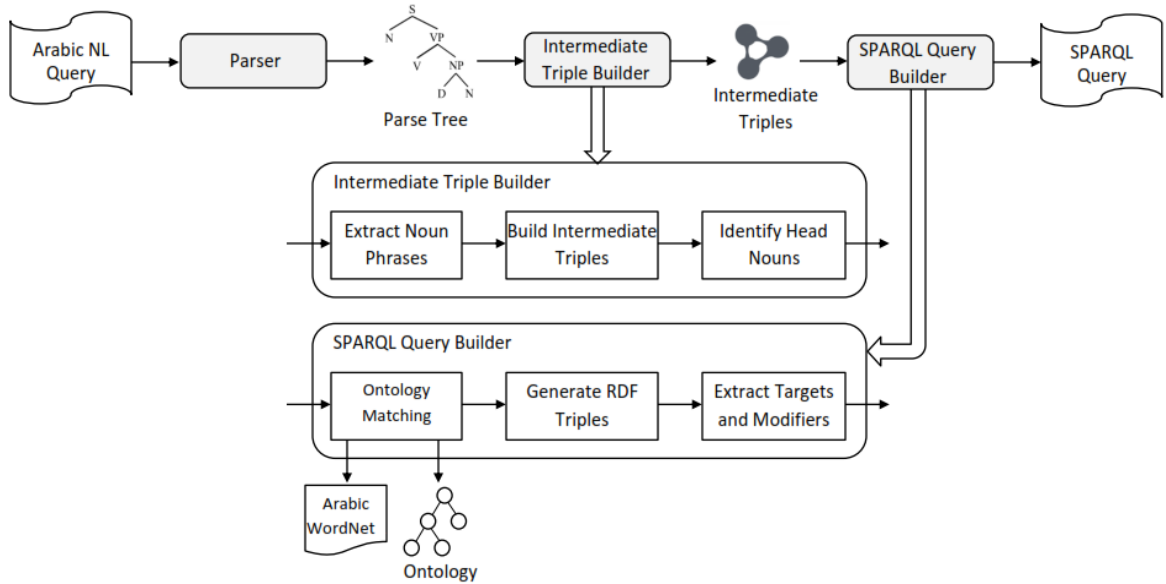

Figure 2. The approach of translating Arabic NL queries to SPARQL

## 4. EXTRACTING NOUN PHRASES FROM PARSE TREES

The idea of translating an Arabic query to SPARQL is based on extracting Noun Phrases (NPs) from Arabic text and then mapping them to RDF triples. A natural language query can be viewed as pairs of NPs that are linked together by using verbs, prepositions, conjunctions or other phrases. Pairs of NPs along with the words linking them are the source of knowledge modelling in ontologies: They can be easily mapped to the RDF triple form <subject, predicate, object> which is the standard format to represent facts in ontologies. The subject and the object of the RDF triple are usually named with NPs and may be classes, instances or literal values. The predicate can be a verb, a verb phrase, a preposition or, sometimes, a noun phrase.

The first step of the translation process is to build a parse tree of the Arabic query from which NPs can be extracted. For this purpose, we used the statistical parser of the Arabic Toolkit Service (ATKS) [6]. ATKS is a set of NLP components proposed by Microsoft and targeting Arabic language. These components have been integrated into several Microsoft services such as Office, Bing, SharePoint and Windows Phone. Recently, all ATKS components have become available for academic use through a web service.

Consider the example query: "ما علاج المرض الذي يسمى داء الملوك؟" whose parse tree is shown in Figure 3". Note that a NP may be a single word, e.g. "علاج" or a combination of words that stand together as a unit, e.g. "داء الملوك". When extracting NPs, it is necessary to identify single-word as well as multi-word NPs to avoid information loss. This can be done as the following: First, single words tagged as nouns are extracted. In the parse tree, a noun is tagged as NN, or any other tag containing NN, e.g. NNS, NNP and DTNN. Referring to Figure 3, the nodes numbered 1, 2, 4 and 5 are extracted. To extract multi-word NPs, we extract NPs whose all leaf nodes are nouns. NPs that only dominate nouns denote complete phrases. For example, the NP numbered 3 in Figure 3 denotes the phrase "داء الملوك". Finally, we end up with the three nodes: 1, 2 and 3. Nodes 4 and 5 are excluded since they are contained in node 3.

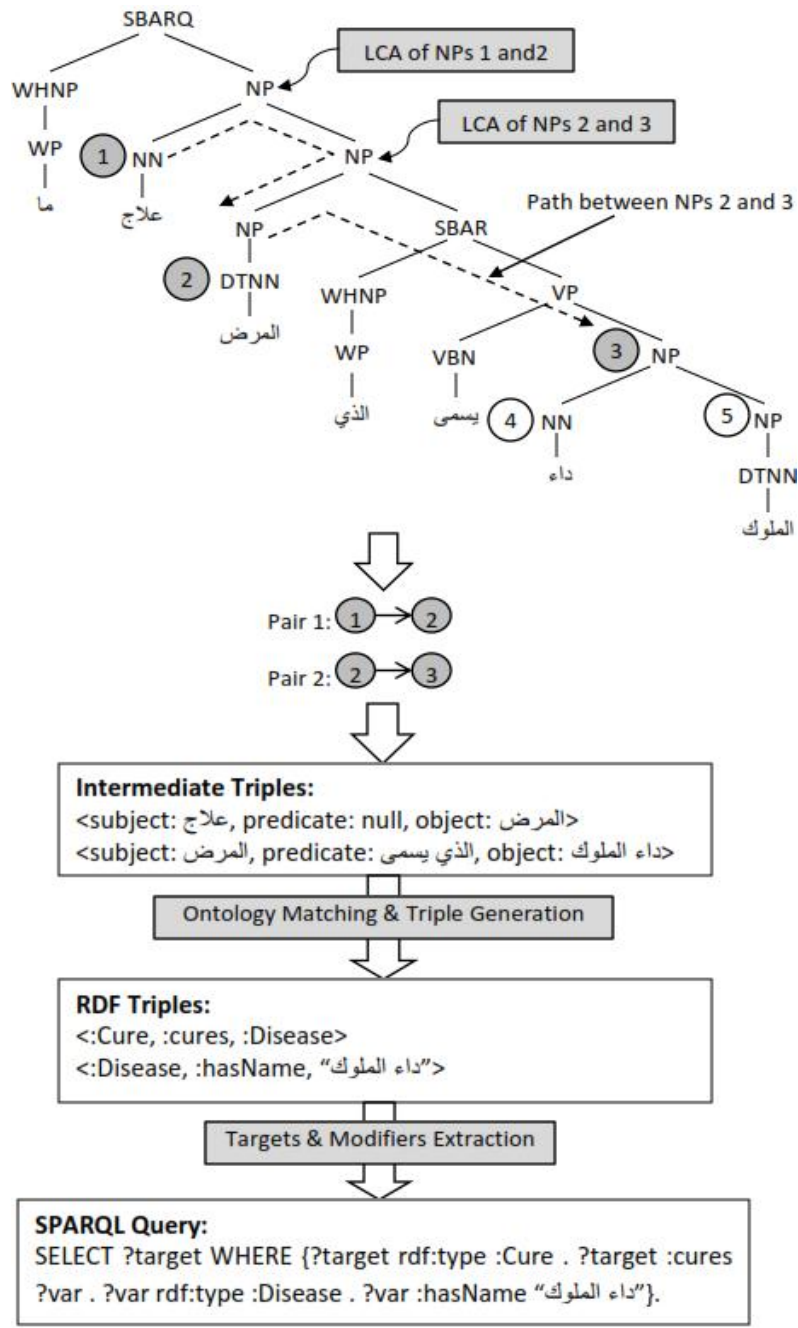

Figure 3: Parse tree of the query: "ما علاج المرض الذي يسمى داء الملوك؟" by the ATKS Parser. The steps of generating the SPARQL query are illustrated.

## 5. EXTRACTING RELATIONS AND BUILDING INTERMEDIATE TRIPLES

After extracting all NPs that are likely to map to ontology terms, the following step is to group these NPs as pairs. Each pair of related NPs corresponds to a candidate RDF triple in the resultant SPARQL query. This is done as the following:

We first order the NPs extracted in the previous step to the order in which they are visited using a pre-order traversal of the parse tree. NP nodes are then grouped into pairs where the second node of the first pair equals the first node of the second pair. Given the parse tree in Figure 3, two pairs are created, which are: <NP1, NP2> and <NP2, NP3>.

Each pair of NP nodes refers to the subject and the object of an RDF triple. Note that to create a complete triple, a relation that links the two NPs should be known. This relation can be determined from the words linking the NP nodes in the parse tree as the following: we find the path between the nodes in each pair. This can be done by finding the lowest common ancestor (LCA) for the two NPs. The LCA for two NPs is the shared ancestor of the nodes that is located farthest from the root (see Figure 3). The two NPs together with the words connecting them through the LCA node are concatenated to form an *Intermediate Triple*. An Intermediate Triple is in the form <subject, predicate, object> and will be translated to an RDF triple in a later phase. The predicate is extracted from the words connecting the two NPs which could be a noun, a verb or a preposition. In the example shown in Figure 3, the following Intermediate Triples are generated by finding the LCAs of NPs and linking them:

- Intermediate Triple 1: <subject: علاج, predicate: null, object: المرض>
- Intermediate Triple 2: <subject: المرض, predicate: الذي يسمى, object: داء الملوك>

Note that we get a null predicate in the first Intermediate Triple because there are no words on the path from NP1 and NP2 (see Figure 3). Missing parts of triples will be identified in a later phase.

The only exception to the above approach is when NPs are children of a Conjunctive Head. A Conjunctive Head is a node containing a word tagged as a conjunction, i.e. "CC". For example, Figure 4 shows the parse tree of the query:"ما الأمراض التي تصيب القلب وتسبب ارتفاع ضغط الدم؟". Note that the NPs 2 and 3 in Figure 4 are linked with a Conjunctive Head. In this case, we ignore the path linking between the children of the Conjunctive Head, e.g. the path linking nodes 2 and 3. This makes sense since the conjunctions "و" and "أو" often link independent clauses. Instead, we consider all the paths linking the preceding, or succeeding, upper-level NP with each child of the Conjunctive Head. In Figure 4, we consider the path linking NP1 with NP2, and the path linking NP1 with NP3. This will generate the following Intermediate Triples:

- Intermediate Triple 1: <subject: الأمراض, predicate: التي تصيب, object: القلب>
- Intermediate Triple 2: <subject: الأمراض, predicate: التي تسبب, object: ارتفاع ضغط الدم>

## 6. IDENTIFYING HEAD NOUNS

Each NP extracted in the previous phase may contain multiple words. Some of these words are essential as they determine the basic meaning of the phrase. These words are often referred as the head of the phrase. Other words may be the head's dependents which modify the head. For example, in the query: "ما أكثر الأمراض المعدية إنتشاراً؟", the noun "الأمراض" is the head noun, while the words "أكثر" and "المعدية" modify the meaning. Head nouns are often mapped to entities in the ontology while non-head nouns can be translated to SPARQL modifiers (e.g. projection, distinct, order by, limit). Therefore, it is necessary to properly capture head and non-head nouns to ensure a valid construction of SPARQL queries.

In this work we only focus on extracting adjectival modifiers [33] which precede or follow the nouns that they modify. Adjectival modifiers include adjectives and other modifiers such as relative clauses and prepositional phrases. For example, the phrase "المرض المعدي" has the adjective "المعدي" as a modifier. The phrase "مدينة في القاهرة" has the prepositional phrase "في القاهرة" as a modifier. These modifiers can be easily extracted from the parse tree by inspecting the POS tags of words preceding or following the NPs. For example, if a two-word phrase starts with a definite noun followed by a definite adjective, e.g. "المرض المعدي", then the first word is considered to be the head noun while the second is a modifier.

After extracting modifiers from NPs, the subject and the object of the Intermediate Triple are represented in the form <pre-modifier . head . post-modifier> where the head noun is the only mandatory part while the modifiers are optional. For example, the phrase "أكثر الأمراض المعدية" is represented as < أكثر (pre-modifier), الأمراض (head), المعدية (post-modifier)>.

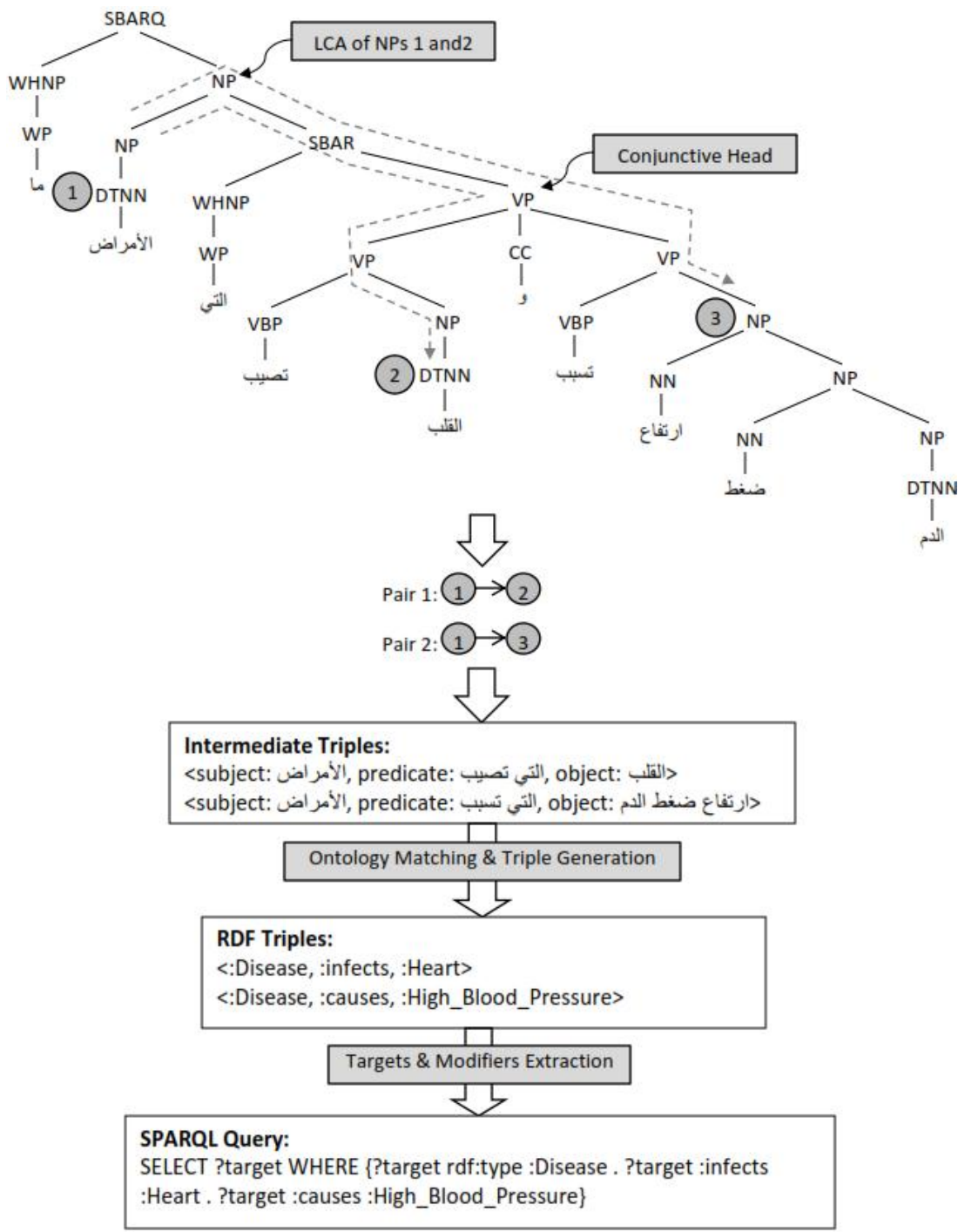

Figure 4: Parse tree of the query: "ما الأمراض التي تصيب القلب وتسبب ارتفاع ضغط الدم؟" by the ATKS Parser. The steps of generating the SPARQL query are illustrated.

## 7. ONTOLOGY MATCHING

In the previous steps, we discussed how to extract NPs from a user query and then group them into pairs to form what we term Intermediate Triples. We also explained how to identify head and non-head nouns of Arabic NPs. The following step is to transform the generated Intermediate Triples to formal RDF triples by matching them with the ontology content. The matching process is done as the following: 1) The heads of NP pairs, which correspond to the subject and the object of the triple, are matched with the ontology classes and instances. The words connecting the NPs, which correspond to the predicate of the triple, are matched with the ontology properties. Matched ontology terms are retrieved and are used to construct the RDF triples.

The content of Intermediate Triples is pre-processed prior to the matching process by applying the following NLP processes (Microsoft ATKS was used):

- Orthographic normalization (e.g. replacing "أ" with "ا" and "ه" with "ة").
- Removal of stop-words and special characters such as "_" that often occurs in ontology text.
- Light Stemming, which aims to make the Arabic words comparable regardless of the different formats.

These pre-processing steps allow for mapping query words to relevant ontology terms even if they are written in different formats.

Note that the query words and the ontology terms may have same meanings but using different formats, synsets or synonyms. For example, the query may contain the word "علاج" while the ontology contains only the word "دواء". To reduce the gap between the user's terminology and the ontology, we used Arabic WordNet (AWN) to find synonyms of query words. AWN is a lexical database, which is structured along the same structures as the Euro WordNet [4] and Princeton WordNet [5, 8].The current implementation of AWN offers an interface to search for Arabic words and retrieve synsets. We integrated AWN into our system so that it is used to find all synonyms of each query word before matching them with the ontology.

To speed up the matching process, we used an Ontological Dictionary which is a special data structure constructed once when the application is first started. It retrieves and stores the whole ontology statements to enable for rapid access and match with the semantic content. Given a word from the user query, the Ontological Dictionary should return matching ontology terms. When the Ontological Dictionary is constructed, we operate an inference engine, i.e. reasoner, to infer additional facts and expressive features based on the given ontology and instance data. This enables the declaration of derived classes or the declaration of further property characteristics (e.g. transitivity and symmetry of properties) which can improve the matching results.

## 8. GENERATING RDF TRIPLES

Having mapped the Intermediate Triples to the ontology content, the following step is to translate the Intermediate Triples to RDF Triples. Resultant RDF Triples should make the body of the target SPARQL query. Ideally, mapping the Intermediate Triples to the ontology content should result in a triple that is compatible with some statements in the ontology in the form <subject, predicate, object>. In this case, the interpretation of the Intermediate Triple into an RDF Triple should be straightforward, and it is done by replacing the subject, the predicate and the object of the Intermediate Triple with their corresponding ontology terms. For example, the Intermediate Triple < داء الملوك, الذي يسمى, المرض> is mapped to the triple <:Disease, :hasName, "داء الملوك">, where :hasName is a data-type property whose literal value is "داء الملوك". Figures 3 and 4 illustrate how Intermediate Triples are converted to RDF triples after the matching process.

The direct conversion of Intermediate Triples to RDF ones may not be always possible. This happens when the Intermediate Triple generated from the parse tree is incomplete, i.e. one or more of the triple parts are missing. In the example shown in Figure 2, the Intermediate Triple <subject: علاج, predicate: null, object: المرض> has no predicate since no word in the NL query could match with a valid predicate.

When a part of the Intermediate Triple is missing, it is possible to utilize the ontology semantics to replace the unknown part with the relevant ontology term. The idea is that if any two parts of the triple are successfully mapped to ontology terms, the third part can be uncovered by capturing ontology statements that best correspond to the incomplete triple. To illustrate how this can be achieved, consider the Intermediate Triple: <subject: علاج, predicate: null, object: المرض>: Matching the Triple with the Diseases ontology gives the following output: <:Cure (Class), null (Predicate), :Disease(Class)>. By knowing both the subject and the object of the triple, the missing predicate can be captured by looking for ontology statements that share the same subject and object. The statement <:Cure, :cures, :Disease> fulfils this condition, and thus the property :cures is used to replace the missing predicate. If multiple ontology statements matches with a single triple, the user is prompted to select the statement that best matches with his/her needs.

Another common problem is the ambiguity resulting when a single word of the Intermediate Triple matches with multiple ontology terms. This ambiguity should be avoided by ensuring a one-to-map mapping, i.e. each word maps to a single ontology term that best describes it. One way to resolve this problem is by verifying the generated RDF triples: Only generated triples that correspond to valid ontology statements are considered. To illustrate how the triple verification is done, consider the schema shown in Figure 1 and the following Intermediate Triple <:the constituent:الذي يصيب> (object) (البنكرياس, predicate) (الذي يصيب), (subject)المرض> matches with two ontology properties which are: "يصيب" (:infects) and "يصاب بـ" (:infected_by) (Note that the stems are similar). This gives two different RDF statements that are: <:Disease, :infects, :Pancreas> and <:Disease, :infected_by, :Pancreas>. These statements are then validated by referring to the ontology semantics and constraints: The first statement corresponds to a valid ontology statement since the subject and the object fall in the domain and the range of the property "يصيب" (:infects) respectively. However, the latter statement does not refer to a valid ontology statement because the property "يصاب بـ" (:infected_by) cannot link between the given subject and object. If multiple statements are found to be valid, the system should prompt the user with a dialog to choose the statement that suits his/her needs.

## 9. IDENTIFYING TARGETS AND MODIFIERS

The SPARQL query typically consists of the parts: the SELECT clause, the WHERE clause and the solution modifiers. The RDF triples generated from the previous steps will be combined together to form the WHERE clause of the resultant SPARQL query. It is still necessary to build the SELECT clause and the solution modifiers, and link them with the WHERE clause in order to have a complete SPARQL query.

To build the SELECT clause, we must identify the targets, i.e. the words that correspond to variables after the "SELECT" word, from the parse tree. This is done as the following: the question words, e.g. "ما","من", are identified. Question words often come at the beginning of the question and are tagged as "WP" in the parse tree. The nominal words in the same or the directly following constituent are extracted as targets. Note that the question may start with an order, e.g. "أذكر, عدد". Therefore, we defined a list of order words and treated them exactly as question words. For example, the targets in the queries illustrated in Figures 3 and 4 are the words: "علاج" and "الأمراض" respectively.

After extracting the targets, we link their corresponding ontology terms with the WHERE clause as the following: 1) we add a variable, e.g. ?target to the SELECT clause. The WHERE clause is then modified by replacing all the occurrences of the target's ontology term :OntTerm with the variable ?target. If the target refers to an ontology class, we add the following triple <?target, rdf:type , :OntTerm> to the WHERE clause. 2) Any term in the WHERE clause that refers to an ontology class is replaced with a variable with an arbitrary name. We then add a rdf:type triple to the WHERE clause with the variable as a subject and the ontology class as an object.

To illustrate the above procedure, consider the parse tree shown in Figure 3. After extracting the Intermediate Triples and mapping them to the ontology we get the following RDF triples:

RDF Triple 1: <:Cure, :cures, :Disease>

RDF Triple 2: <:Disease, :hasName, "داء الملوك">

The target of the query is the word "علاج" as it directly follows the question word. The ontology class that corresponds to this noun is :Cure. Accordingly, a variable with an arbitrary name, i.e. ?target, is added to the SELECT clause. Then, all the occurrences of the class :Cure are replaced with the variable ?target. As the term :Disease refers to an ontology class, it is also replaced with a variable, e.g. ?var. The RDF triple <?var, rdf:type, :Disease> is also added to the WHERE clause. This results in the following SPARQL query:

SELECT ?target WHERE {?target rdf:type :Cure . ?target :cures ?var . ?var rdf:type :Disease . ?var :hasName "داء الملوك"}.

The *FILTER* clause is used within the curly parenthesis as a sub-clause of the WHERE clause. As its name suggests, it enables for filtering the query results based on specific conditions. Solution modifiers are optionally used to apply some operations, e.g. order, projection, limit, on the query results.

It is necessary first to find out whether there is a need for a FILTER clause or a solution modifier. This can be determined by looking for query words that refer to these modifiers. In the current version of our approach, we mainly look for the following types of words which we called modifier descriptors: 1) Negation: denoted by the words "لا" and "غير". 2) conjunctive/disjunctive, including "و" and "أو". These modifier descriptors are all extracted from the user query along with their types and positions in order to be considered while generating the SPARQL query. Our approach does not currently handle comparative and superlative words such as "أبرز/أهم/المشابه لـ/أكبر" (main, most, largest) since the interpretation of these modifiers often requires special techniques to understand the comparison in different ontologies. For example, in the query "ما أهم الأمراض ..؟", it is unclear how the importance of a disease is evaluated. However, domain-specific rules can be defined later to support the interpretation of superlative/comparative modifiers.

Extracted modifier descriptors are interpreted as the following:

1. Negations on the property are interpreted by using both "OPTIONAL" and "FILTER" clauses. For example, in the query "ما الأمراض التي لا تعالج بالمضادات الحيوية؟", the words " لا تعالج" is interpreted as "OPTIONAL {{?disease  :cured_by ?cure} FILTER(?cure = :Antibiotics)} FILTER(!bound(?cure))".
2. Conjunctive/disjunctive modifiers: RDF triples linked with "و" are interpreted as conjunctive triple patterns to the WHERE clause by default. Triples linked with "or" are interpreted with a linking "UNION". For example, in the query "ما الأمراض الذي تصيب القلب أو الرئتين؟", two triples are linked with UNION as the following: :SELECT ?disease WHERE {{ ?disease :infects :heart} UNION {?disease :infects :Lung}}.

## 10. EVALUATION

The objective of the evaluation is to quantitatively assess the ability to translate Arabic NL queries to valid SPARQL queries that, when executed, adequate answers will be retrieved from an ontology-based knowledge base.

To achieve that, we built a prototype desktop application that allows the user, through a simple user interface, to input the NL query expressed in Arabic and then activates the translation process. The current implementation uses Jena API to access and process the underlying ontology. It relies on the Arabic NLP Toolkit Service (ATKS), from Microsoft, for constructing parse trees of Arabic text besides other NLP processes. The application also integrates the Arabic WordNet (AWN) to extend the user's terminology when matching the query text with the ontology terms.

### 10.1. Datasets

In the domain of English language, NL interfaces to ontologies have been often evaluated by using widely-used OWL test data and queries such as [34]. However, we are unaware of similar standardized test data for the Arabic language. Therefore, we prepared and used two different datasets for our evaluation: the first dataset is based on the widely-used Mooney's dataset and queries on the geography of the United States. The dataset consists of an OWL ontology and 877 queries expressed in English. The dataset was modified for our evaluation as the following: the ontology was populated with Arabic translations of all ontology classes, properties and instances. Translations were added to the ontology through the rdfs:label property. The queries were also translated to Arabic, and all translations were validated by a professional translator.

The second dataset consists of the Diseases ontology which part of is shown in Figure 1. The ontology was developed and validated with the help of a domain expert. It contains a total of 24 classes, 12 object-type properties and 8 data-type properties. Ontology terms were translated to Arabic, and translations were added to the ontology using the rdfs:label property. A total of 124 instances of different types were created and linked using appropriate relations from the ontology. The query set was created with the help of five human subjects, i.e. medical students, who were familiar with the ontology domain. Each student was asked to formulate 10 different queries. In total, 45 questions were chosen after excluding duplicated ones.

We made the full datasets freely available for academic use through https://goo.gl/9CWcQs. One advantage of testing the system with two datasets was to assess the system's performance and portability when it is interfaced to different ontologies.

### 10.2. Evaluation Metrics

To assess our approach's correctness, the SPARQL queries generated by the approach were compared with the manually generated SPARQL queries for each dataset. We used precision and recall metrics, which are defined as the following:

$$\text{Precision} = \frac{\text{number of correctly translated queries}}{\text{number of queries generated by the system}}$$

$$\text{Recall} = \frac{\text{number of correctly translated queries}}{\text{number of testing queries}}$$

### 10.3. Results and Discussion

Table 1 summarizes the results for the two datasets. Using the Mooney's geographic ontology, the approach successfully handled 514 queries, achieving 80.56% precision and 58.61% recall.

Table 1. Evaluation of the system using the Mooney's geography and Diseases ontologies.

| Ontology | Mooney's Geography | Diseases |
| --- | --- | --- |
| #. of queries | 877 | 45 |
| #. of queries generated by the system | 638 | 39 |
| # of correctly generated queries | 514 | 31 |
| Precision | 80.56% | 79.49% |
| Recall | 58.61% | 68.89% |

Of the remaining queries, 124 queries (14% of the query set) were incorrectly translated, and 239 queries (27% of the query set) were not translated at all. When testing with the Diseases ontology, the system successfully handled 31 queries, achieving 79.49% precision and 68.89% recall. 8 queries (18% of the query set) were incorrectly translated while 14 queries were not translated at all.

The above results show that our approach achieved low recall ratios for both datasets. This can be explained mainly by the many queries that are not currently supported, especially in the Mooney's geography dataset. Most unsupported queries contain comparative/superlative modifiers or require deep inferences that are not currently supported. On excluding unsupported queries from the two query sets, the average recall reaches 78.82% and 76.5% respectively.

In the following, the main sources of errors that we discovered after inspecting results are discussed:

- *Parsing of Arabic text:* This type of errors originated from the incorrect parsing of some Arabic queries. Incorrectly generated parse trees often led to the generation of incorrect Intermediate Triples. A closer analysis shows that parse failures often occurred with: 1) compound sentences consisting of clauses connected by a coordinating conjunction such as "و" and "أو". The ATKS parser sometimes failed to correctly identify the conjunct dependencies especially for long sentences. For example, in the query:"ما الأمراض التي تصيب البنكرياس وتسبب عسر الهضم؟": there is a conjunct relation between the two verb phrases "تصيب البنكرياس" and "تسبب عسر الهضم", and both phrases refer to the same subject "الأمراض". However, the ATKS parser generated a conjunct relation between the noun "البنكرياس" and the phrase "تسبب عسر الهضم", and both refer to the verb "تصيب". 2) Structural ambiguity caused by affixed pronouns and the lack of discretization which caused different possible readings of an input sentence. For example, the system failed to handle the query "ما مسببات الصداف وما أعراضه؟" because it could not replace the pronoun in the word "أعراضه" with the noun "الصداف".

Incorrect parsing of Arabic text was the major source of errors and accounted for 44% and 43% of the total number of errors for the Geography and Diseases datasets respectively. Although the parsing results observed in our experiment seem satisfactory, this result indicates that Arabic text parsing still demands further investigation into the syntactic ambiguities in Arabic sentences. However, enhancing the parsing results is out of the scope of this work.

- *Entity Identification*: Some errors occurred when words could not be mapped to any ontology entity, a thing that resulted in incomplete triples. The most frequent causes of this type of errors were as the following: 1) lack of semantic matching: some words can match with ontology entities semantically but not syntactically. However, our approach is limited to syntactic matching. Consider the following query: "ما النهر الذي يخترق معظم الولايات؟", the word "يخترق" does not syntactically map to any ontology entity even though it semantically corresponds to the property "يمر عبر (:runsThrough)". 2) Implicit relations: some queries do not contain explicit relations that can map to valid ontology properties. For example, in the query "أذكر أسماء المدن في ولاية تكساس؟", the preposition "في" is supposed to be replaced by the ontology property ":isCityOf". However, our approach gave two possible replacements for the preposition "في", which are: ":isCityOf" and ":borders". Determining the correct implicit property needs inferences that are not part of our approach at the moment. In general, this type of errors accounted for 18% and 36% of the total number of failed queries for the Geography and Diseases datasets respectively.

- *Lack of semantic analysis:* Some queries do not have answers that can be directly retrieved from the ontology. Answering these queries requires deep analysis or reasoning to be performed. Examples of failed queries due to this error are: "ما هو الطول الإجمالي لجميع الأنهار في الولايات المتحدة الأمريكية؟" and "ما هي عاصمة الولاية التي يحدها أكبر عدد من الولايات؟". Answers to these queries are not explicitly present in the ontology, and require calculations to be made. In addition, some words can be interpreted differently according to the context. For example, in the query "ما هي المدن الرئيسية في أكبر ولاية؟" ('What are the major cities in the largest state?), it is unclear whether the comparative and superlative words "أكبر, الرئيسية" refer to the area or the population size. This type of errors accounted for 39% and 21% of the total number of failed queries for the Geography and Diseases datasets respectively. Note that this error was more common with the Geography dataset since it contains more analytical questions. In contrast, the queries in the Diseases dataset are mostly straightforward and do not require deep inference.

## 11. CONCLUSION AND FUTURE WORK

We presented an approach to Arabic question answering over ontologies and RDF stores. The approach relies on deep linguistic analysis to construct parse trees of Arabic queries and extract NPs. Pairs of NPs and the words linking them form what we termed "Intermediate Triples". Intermediate Triples are then converted to RDF triples by replacing NPs with the relevant terms from the ontology and by exploiting the ontology content to replace any missing parts of RDF triples. Finally, a SPARQL query is generated after identifying the targets and modifiers of the query.

We believe that the proposed approach makes a step towards enabling naïve Arab users to interact with the growing Arabic content on the Semantic Web. One of the strengths of this approach is that it uses an off-the-shelf Arabic language parser to perform linguistic analysis. This helped to produce results that are comparable with those approaches working on English queries [26, 31].

In our future work, we plan to resolve some of the limitations we explored in our evaluation. We first aim to enhance the reasoning capabilities so that the approach can handle comparative, superlative modifiers and other queries that need deep inference. Second, we aim to improve the ontology matching process to support semantic rather than syntactic matching. This can be achieved by incorporating semantic similarity measures when comparing the query words with the ontology terms. To test the feasibility of our approach in practice, we will apply it to support question answering over popular ontologies that support Arabic, such as the Quranic Ontology [35], and other ontology-based Islamic resources [21, 24]. Finally, we will explore approaches to enhance the parsing results by testing other parsers of Arabic text and comparing results.

**Authors**

**Iyad M. AlAgha** received his MSc and PhD in Computer Science from the University of Durham, the UK. He worked as a research associate in the center of technology enhanced learning at the University of Durham, investigating the use of Multi-touch devices for learning and teaching. He is currently working as an assistant professor at the Faculty of Information technology at the Islamic University of Gaza, Palestine. His research interests are Semantic Web technology, Adaptive Hypermedia, Human-Computer Interaction and Technology Enhanced Learning.

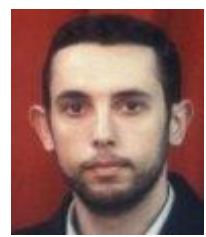